\documentclass[11pt]{memoir}

\usepackage[english]{babel}

\usepackage{caption}
\usepackage[letterpaper,top=2cm,bottom=2cm,left=3cm,right=3cm,marginparwidth=1.75cm]{geometry}

\usepackage{amsmath,bm,bbm,amssymb}
\usepackage{graphicx}
\usepackage{wrapfig}
\usepackage{endnotes}

\def\mysection#1{\section*{{\large #1}}}
\def\mysubsection#1{\section*{{\normalsize #1}}}

\def\x{{\bm x}}
\def\w{{\bm w}}
\def\y{{\bm y}}
\def\tr{{^\top}}

\begin{document}

\begin{center}
{\Large\textbf{The Alberta Plan for AI Research}}\\
~\\
{\textbf{Richard S. Sutton, Michael Bowling, and Patrick M. Pilarski}\\
\vspace{5pt}
University of Alberta\\Alberta Machine Intelligence Institute\\DeepMind Alberta\\
}
\end{center}

\bigskip
\chapterprecishere{\small History suggests that the road to a firm research consensus is extraordinarily arduous.\par\raggedleft--- \textup{Thomas Kuhn}, The Structure of Scientific Revolutions}
\bigskip

Herein we describe our approach to artificial intelligence (AI) research, which we call \emph{the Alberta Plan}. The Alberta Plan is pursued within our research groups in Alberta and by others who are like minded throughout the world. We welcome all who would join us in this pursuit.

The Alberta Plan is a long-term plan oriented toward basic understanding of computational intelligence. It is a plan for the next 5--10 years. It is not concerned with immediate applications of what we currently know how to do, but rather with filling in the gaps in our current understanding. As computational intelligence comes to be understood it will undoubtedly profoundly affect our economy, our society, and our individual lives. Although all the consequences are difficult to foresee, and every powerful technology contains the potential for abuse, we are convinced that the existence of more far-sighted and complex intelligence will overall be good for the world.

Following the Alberta Plan, we seek to understand and create long-lived computational agents that interact with a vastly more complex world and come to predict and control their sensory input signals. The agents are complex only because they interact with a complex world over a long period of time; their initial design is as simple, general, and scalable as possible. To control their input signals, the agents must take action. To adapt to change and the complexity of the world, they must continually learn. To adapt rapidly, they must plan with a learned model of the world.

The purpose of this document is twofold. One is to describe our vision for AI research and its underlying intellectual commitments and priorities. 
The second is to describe the path along which this vision may unfold and the research problems and projects that we will pursue. What we say toward the first goal is described here in order to establish a clear record of where we come from; this part of our research strategy is expected to be relatively stable. What we say toward the second goal is much more contemporary. The voyage is uncertain; our path has gaps and uncertainties. Nevertheless, we attempt to chart the path ahead with as much specificity as possible so that we know where we are trying to go, even if in the end we go another way or arrive at a somewhat different destination.

\mysection{Research Vision: Intelligence as signal processing over time}

\begin{figure}[t]
\centerline{\includegraphics[width=0.50\textwidth]{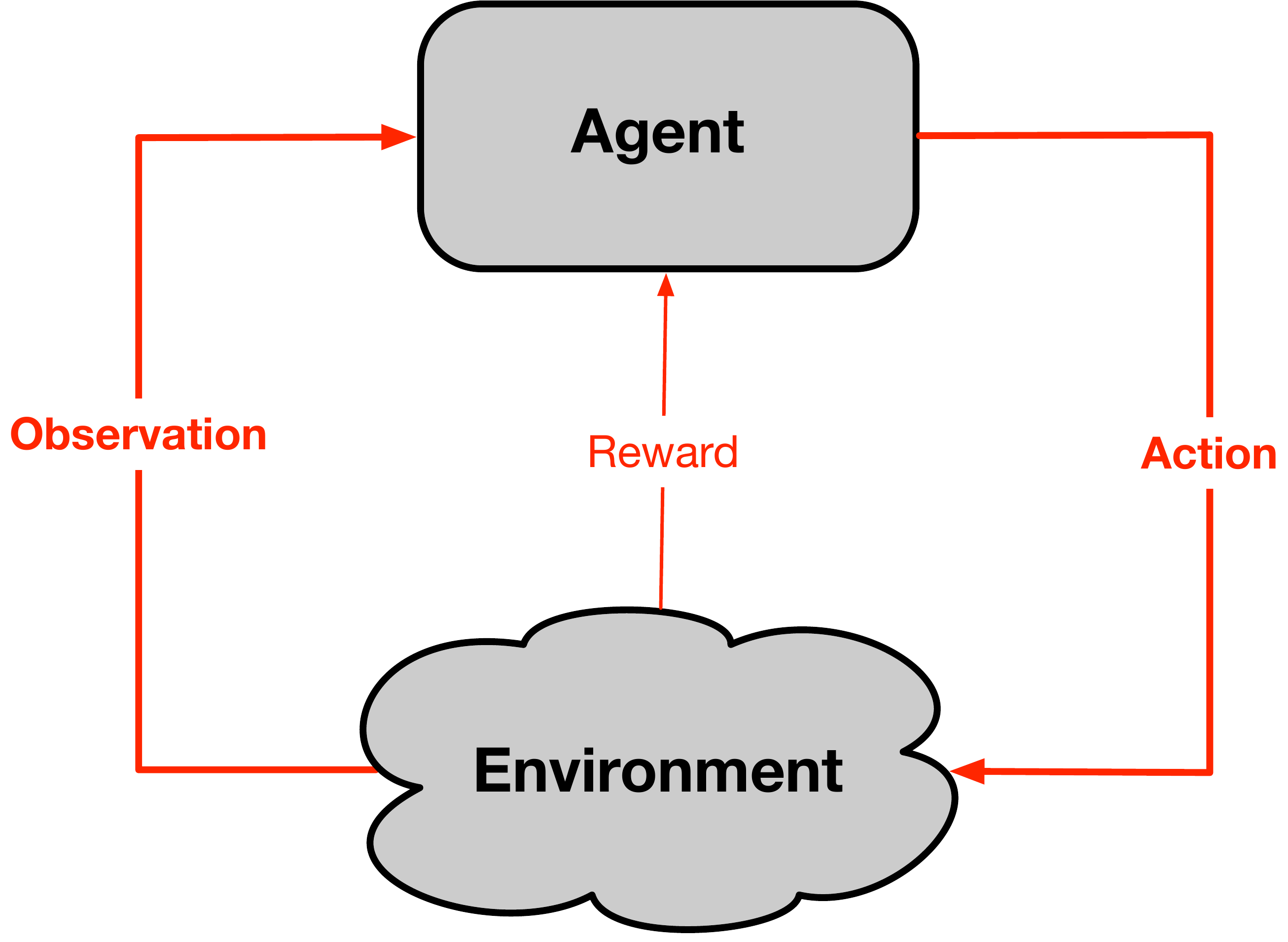}}
\caption{In the Alberta Plan's research vision, an intelligent agent receives observation and reward signals from its environment and seeks to control those signals with its actions. 
This is the standard perspective in advanced reinforcement learning.}
\end{figure}

We seek to understand and create long-lived computational agents that interact with a vastly more complex world and come to \emph{predict and control their sensory input signals}, particularly a distinguished scalar signal called reward. 
The overall setting we consider is familiar from the field of reinforcement learning\endnote{E.g., see Sutton \& Barto (2018).} (Figure 1).
An agent and an environment exchange signals on a fine time scale. The agent sends \textit{actions} to the environment and receives sensory signals back from it. The larger sensory signal, the \textit{observation}, is explicitly \emph{not} expected to provide complete information about the state of the environment.\endnote{This environment is technically a Partially Observable Markov Decision Process (POMDP, see, e.g., Lovejoy 1991), but that formalism is of little use here.}
The second sensory signal, the \textit{reward}, is scalar and defines the ultimate goal of the agent---to maximize the total reward summed over time. 
These three time series---\emph{observation, action, and reward}---constitute the \textit{experience} of the agent. We expect all learning to be grounded in these three signals and not in variables internal to the environment. Only experience is available to the agent, and the environment is known only as a source and sink for these signals.

The first distinguishing feature of the Alberta Plan's research vision is its emphasis on ordinary experience, as described above, as opposed to special training sets, human assistance, or access to the internal structure of the world. Although there are many ways human input and domain knowledge can be used to improve the performance of an AI, such methods typically do not scale with computational resources and as such are not a research priority for us.\endnote{The limits of such approaches are described accessibly in the blog post \emph{The Bitter Lesson} (Sutton 2019) and the talk \emph{The Future of Artificial Intelligence Belongs to Search and Learning} (Sutton 2016).}

The second distinguishing feature of the Alberta Plan's research vision can be summarized in the phrase \emph{temporal uniformity}. Temporal uniformity means that all times are the same with respect to the algorithms running on the agent. There are no special training periods when training information is available or when rewards are counted more or less than others. If training information is provided, as it is via the reward signal, then it is provided on every time step. If the agent learns or plans, then it learns or plans on every time step. If the agent constructs its own representations or subtasks, then the meta-algorithms for constructing them operate on every time step. If the agent can reduce its speed of learning about parts of the environment when they appear stable, then it can also increase its speed of learning when they start to change. Our focus on temporally uniform problems and algorithms leads us to interest in non-stationary, continuing environments and in algorithms for continual learning and meta-learning.\endnote{The interplay between non-stationary environments, continual learning, and meta-learning has begun to be explored in a variety of works. Useful surveys are provided by Hadsell et al.\ (2020), Parisi et al.\ (2019), and Khetarpal et al.\ (2020).
We also recommend the works by Thrun \& Pratt (1998), Sutton, Koop \& Silver (2004), Sutton (1991), Dohare (2020),  Rahman (2020), and Samani (2022).}

Temporal uniformity is partly a constraint on what we research and partly a discipline that we impose on ourselves. Keeping everything temporally uniform reduces degrees of freedom and shrinks the agent-design space. Why not keep everything temporally uniform? Having posed that rhetorical question, we acknowledge that there may be situations in which it is preferable to depart from absolute temporal uniformity. But when we do so, we are aware that we are stepping outside this discipline.

\newpage
The third distinguishing feature of the Alberta Plan research vision is its cognizance of computational considerations. Moore's law and its generalizations bring steady exponential increases in computer power, and we must prioritize methods that scale proportionally to that computer power. 
Computer power, though exponentially more plentiful, is never infinite. The more we have, the more important it is to use it efficiently, because it is a greater and greater determinant of our agents' performance. We must heed the bitter lesson of AI's past and prioritize methods, such as learning and search, that scale extensively with computer power, while de-emphasizing methods that do not, such as human insight into the problem domain and human-labeled training sets.$^3$

Beyond these large-scale implications, computational considerations enter into every aspect of an intelligent agent's design. For example, it is generally important for an intelligent agent to be able to react quickly to a change in its observation. But, given the computational limitations there is always a tradeoff between reaction time and the quality of the decision. The time steps should be of uniform length. If we want the agent to respond quickly, then the time step must be small---smaller than would be needed to identify the best action. A better action might be available from planning, but planning, and even learning, takes time; sometimes it is better to act fast than to act well.

Giving priority to reactive action in this way does not preclude an important role for planning. The reactive policy may recommend a temporizing action until planning has improved the policy before a more committal action is taken, just as a chess player may wait until she is sure of her move before making it. Planning is an essential part of intelligence and our or research vision.

The fourth distinguishing feature of the Alberta Plan research vision is that it includes a focus on the special case in which the environment includes other intelligent agents. In this case the primary agent may learn to communicate, cooperate, and compete with the environment and should be cognizant that the environment may behave differently \emph{in response} to its action. AI research into game playing must often deal with these issues.
The case of two or more cooperating agents also includes cognitive assistants and prostheses. This case is studied as \emph{Intelligence Amplification} (IA), 
a subfield of human-machine interaction.\endnote{The phrase "Intelligence Amplification" and its role as a subfield of study was first established in the 1950s and 1960s, notably by Ashby (1956), and in subsequent work by Licklider (1960) and Engelbart (1962).} There are general principles by which one agent may use what it learns to amplify and enhance the action, perception, and cognition of another agent, and this amplification is an important part of attaining the full potential of AI. 

The Alberta Plan characterizes the problem of AI as the online maximization of reward via continual sensing and acting, with limited computation, and potentially in the presence of other agents. This characterization might seem natural, even obvious, but it is also contrary to current practice, which is often focused on offline learning, prepared training sets, human assistance, and unlimited computation. The Alberta Plan research vision is both classical and contrarian, and radical in the sense of going to the root.

\mysection{Research Plan}

All research plans are suspect and provisional. Nevertheless, we must make them in order to communicate among ourselves and collaborate efficiently. The Alberta Plan is not meant to be a limit on what members of our teams do individually, but an attempt at consensus on what we do together.

\mysubsection{Designing around a base agent}

\begin{figure}[b]
~~~~~~~~~~~~~~~\includegraphics[width=0.65\textwidth]{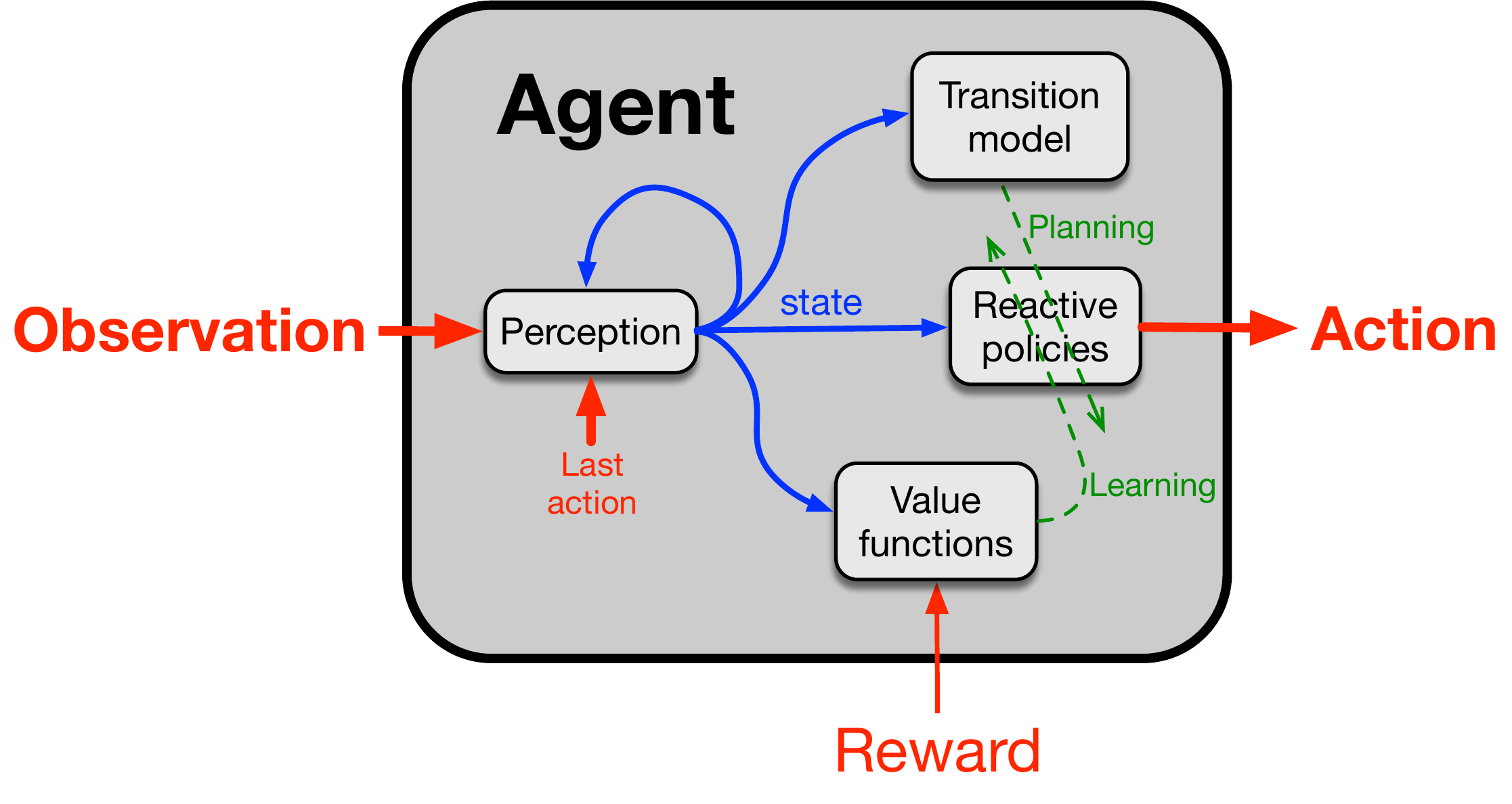}
\caption{\small The base agent of the Alberta Plan consists of four components interconnected by a state signal constructed by the perception component. All components may be learned.}
\end{figure}

Our research in agent design begins with the standard or base agent shown in Figure 2, which is itself based on the ``Common Model of the Intelligent Agent" that has been proposed as common to AI, psychology, control theory, neuroscience, and economics (Sutton 2022). Our base agent has four primary internal components. \emph{Perception} is the component that updates the agent's summary of its past experience, or \emph{state}, which is then used by all components. The \emph{reactive policies} component includes the \emph{primary policy}, which selects the action that will be sent to the environment and which will be updated toward the goal of maximizing reward. Perception and the primary policy together map observations to actions and thus can serve as a minimal agent. Our base agent allows for the possibility of other reactive policies, perhaps maximizing quantities other than reward. Each policy has a corresponding \emph{value function} that is used to learn it. The set of all value functions form the \emph{value functions} component. Allowing multiple policies and value functions is the main way our base agent differs from the Common Model of the Intelligent Agent.

The fourth component of the base agent, the \emph{transition model} component, represents the agent's knowledge of the world's dynamics. The transition model is learned from observed actions, rewards, and states, without involving the observations. Once learned, the transition model can take a state and an action and predict the next state and the next reward. In general, the model may be temporally abstract, meaning that it takes not an action, but an \emph{option} (a policy together with a termination condition),\endnote{For more on options see the works by Sutton, Precup \& Singh (1999), Precup (2000), and Sutton \& Barto (2018, Section 7.2).}
and predicts the state at the time the option terminates and the cumulative reward along the way. The transition model is used to imagine possible outcomes of taking the action/option, which are then evaluated by the value functions to change the policies and the value functions themselves. This process is called \emph{planning}. Planning, like everything else in the architecture, is expected to be continual and temporally uniform. On every step there will be some amount of planning, perhaps a series of small planning steps, but planning would typically not be complete in a single time step and thus would be slow compared to the speed of agent--environment interaction.

Planning is an ongoing process that operates asynchronously, in the \emph{background}, whenever it can be done without interfering with the first three components, all of which must operate on every time step and are said to run in the \emph{foreground}.\endnote{As in the work of Daniel Kahneman (2011) on Systems 1 and 2.}
 On every step the new observation must be processed by perception to produce a state, which is then processed by the primary policy to produce that time step's action. The value functions must also operate in the foreground to evaluate each time step's new state and the decision to take the previous action. Our strong preference is to fully process events as they occur. In particular, all four components are updated by learning processes operating in the foreground using the most recent events together with short-term credit-assignment memories such as eligibility traces.\endnote{See Sutton and Barto (2018), Chapter 13.}

Our base agent is a starting point from which we often deviate or extend. The perception component is perhaps the least well understood. Although we have examples of static, designed perception processes (such as belief-state updating or remembering four frames in Atari), how perception should be learned, or meta-learned, to maximally support the other components remains an open research question. Planning similarly has well understood instantiations, and yet how to do it effectively and generally---with approximation, temporal abstraction, and stochasticity---remains unclear. The base agent also does not include \emph{subtasks}, even though these may be key to the discovery of useful options.\endnote{See, e.g., Jaderberg et al.\ (2016), Sutton, et al.\  (2022).} Also unmentioned in the base agent are algorithms to direct the planning process, such as prioritized sweeping,\endnote{See Moore \& Atkeson (1993), Peng \& Williams (1993).} sometimes referred to generically as \emph{search control}. Perhaps the best understood parts of the base agent are the learning algorithms for  the value functions and reactive policies, but even here there is room for improvement in their advanced forms, such as those involving average reward, off-policy learning, and continual non-linear learning.$^1$ Finally, the learning of the world model, given the options, is conceptually clear but remains challenging and underexplored. Better understanding of advanced forms of all of these algorithms are important areas for further research. Some of these are discussed further in the next section.

\mysubsection{Roadmap to an AI Prototype}

The word ``roadmap" suggests the charting of a linear path, a sequence of steps that should be taken and passed through in order. This is not completely wrong, but it fails to recognize the uncertainties and opportunities of research. The steps we outline below have multiple interdependencies beyond those flowing from first to last. The roadmap suggests an ordering that is natural, but which will often be departed from in practice. Useful research can be done by entering at or attaching to any step. As one example, many of us have recently made interesting progress on integrated architectures even though these appear only in the last steps in the ordering.

\newcounter{item}
\def\myitem#1#2{\addtocounter{item}{1}\newline\hspace*{.2in}\arabic{item}. #1: #2.}

First let's try to obtain an overall sense of the roadmap and its rationale. There are twelve steps, titled as follows:
\smallskip
\myitem{Representation I}{Continual supervised learning with given features} 
\myitem{Representation II}{Supervised feature finding}
\myitem{Prediction I}{Continual Generalized Value Function (GVF) prediction learning} 
\myitem{Control I}{Continual actor-critic control}
\myitem{Prediction II}{Average-reward GVF learning}
\myitem{Control II}{Continuing control problems}
\myitem{Planning I}{Planning with average reward}
\myitem{Prototype-AI I}{One-step model-based RL with continual function approximation}
\myitem{Planning II}{Search control and exploration}
\myitem{Prototype-AI II}{The STOMP progression}
\myitem{Prototype-AI III}{Oak}
\myitem{Prototype-IA}{Intelligence amplification}
\smallskip\newline
The steps progress from the development of novel algorithms for core abilities (for representation, prediction, planning, and control) toward the combination of those algorithms to produce complete prototype systems for continual, model-based AI. 

\newpage
An eternal dilemma in AI is that of ``the parts and the whole."
A complete AI system cannot be built until effective algorithms for the core abilities exist, but exactly which core abilities are required cannot be known until a complete system has been assembled. To solve this chicken-and-egg problem, we must work on both chickens and eggs, systems and component algorithms, parts and wholes, in parallel. The result is imperfect, with wasted effort, but probably unavoidably so.


The idea of this ordering is to front load---to encounter the challenging issues as early as possible so that they can be worked out first in their simplest possible setting.


\setcounter{item}{0}
\def\myitem#1#2{\addtocounter{item}{1}\item[Step \arabic{item}. #1: #2.]}

\begin{description}
\myitem{Representation I}{Continual supervised learning with given features} 
Step 1 is exemplary of the primary strategy of the Alberta Plan: to focus on a particular issue by considering the \emph{simplest setting} in which it arises and attempting to deal with it there, fully, before generalizing to more complex settings. 
The issues focussed on in Step 1 are continual learning and the meta-learning of representations. How can learning be rapid, robust, and efficient while continuing over long periods of time? How can long periods of learning be taken advantage of to meta-learn better representations, and thus to learn most efficiently? 

The simple setting employed in Step 1 is that of supervised learning and stochastic gradient descent with a linear function approximator with static, given features. In this setting, conventional stochastic-gradient-descent methods such as the Least-Mean-Square learning rule work reasonably well even if the problem is non-stationary. However, these methods can be significantly improved in their efficiency and robustness, and that is the purpose of Step 1. First, these methods often involve a global step-size parameter that must be set by an expert user aided by knowledge of the target outputs, the features, the number of features, and heuristics. All of that user expertise should be replaced by a meta-algorithm for setting the step-size parameter so that the same method can be used on any problem or for any part of a large problem. Second, instead of a global step-size parameter, there should be different step-size parameters for each feature depending on how much generalization should be done along that feature. If this is done, then there will be many step-size parameters to set, and it will be even more important to set them algorithmically.

In this setting, the representations are the features, which are given and fixed, so it may seem surprising to offer the setting up as a way of exploring representation learning. It is true that the setting cannot be used to discover features or to search for new features, but it \emph{can} be used to assess the utility of given features---an important precursor to full representation discovery. Even without changing the features it is possible to learn which features are relevant and which are not. The relevant features can be given large step-size parameters and the irrelevant features small ones; this itself is a kind of representation learning than can affect learning efficiency even without changing the features themselves.

Finally, there are normalizations of the features (scalings and offsets) that can greatly affect learning efficiency without changing the representational power of the linear function approximator, and we include these in Step 1.

In particular, we consider an infinite sequence of examples of desired behavior, each consisting of a real-valued input vector paired with a real-valued desired output. Let the $t$th example be a pair denoted $(\x_t, y_t^*)$. The learner seeks to find an affine mapping from each input vector $\x_t$ to an output $y_t$ that closely approximates the desired output $y_t^*$. That affine map is represented as a vector of weights $\w_t$ and a scalar bias or offset term $b_t$. That is, the output is $y_t\doteq \w_t\tr\x_t + b_t$. The objective is to minimize the squared error $(y_t^*-y_t)^2$ by learning $\w_t$ and $b_t$. Each example is independent, but the distribution from which it is generated changes over time, making the problem non-stationary. In particular, we can take the desired output as being affine in the input vector and an unknown target weight vector $\w_t^*$ that changes slowly over time, plus an additional, independent mean-zero noise signal: $y_t^* \doteq \w_t^*\tr\x_t + b_t^* + \eta_t$. The problem is non-stationary if $\w_t^*$ or $b_t^*$ change over time or if the distribution of $\x_t$ changes over time.

In this simple setting, there are still essential questions that have yet to be definitively answered. We are particularly interested in questions of \emph{normalization} and \emph{step-size adaptation}. Without changing the expressive power of our linear learning unit or the order of its computational complexity, we can transform the individual inputs $x_t^i$ to produce normalized signals $\tilde x_t^i\doteq \frac{x_t^i-\mu_t^i}{\sigma_t^i}$ where $\mu_t^i$ and $\sigma_t^i$ are non-stationary (tracking) estimates of the $i$th signal's mean and standard deviation. Surprisingly, the effect of such online normalization has yet to be definitively established in the literature. We consider learning rules of the form:
\begin{equation}
w_{t+1}^i \doteq w_t^i + \alpha_t^i \left(y_t^*-y_t\right) \tilde x_t^i, ~~\forall i,  
\end{equation}
where each $\alpha_t^i$ is a meta-learned, per-weight, step-size parameter, and 
\begin{equation}
b_{t+1} \doteq b_t + \alpha_t^b \left(y_t^*-b_t\right),   
\end{equation}
where $\alpha_t^b$ is another potentially-meta-learned step-size parameter.
Our initial studies for Step 1 will focus on algorithms for meta-learning the step-size parameters, building on existing algorithms,\endnote{Existing algorithms include NADALINE (Sutton 1988b), IDBD (Sutton 1992a,b), Autostep (Mahmood et al.\ 2012), Autostep for GTD($\lambda$) (Kearney et al.\ 2022), Auto (Degris in prep.), Adam (Kingma \& Ba 2014), RMSprop (Tieleman \& Hinton 2012), and Batch Normalization (e.g., Ioffe \& Szegedy 2015).}
and on demonstrating their improved robustness.

The overall idea of Step 1 is to design as powerful an algorithm as possible given a fixed feature representation. It should include all the most important issues of non-stationarity in the problem (for a fixed set of linear features), including the tracking of changes in feature relevance.
It should include the meta-learning of feature relevance, a challenging issue in representation learning---arguably the most challenging issue---but it does not include actually changing the set of features under consideration; that is explored in Step 2. 

\myitem{Representation II}{Supervised feature finding}

This step is focused on creating and introducing new features (made by combining existing features) in the context of continual supervised learning as in Step 1, except now targets will be vectors $\y^*_t$ approximated by output vectors $\y_t$. Getting each component of $\y_t$ to match $\y^*_t$ is referred to as a separate task. How should new features be constructed from the existing ones to maximize the new features potential utility and the speed with which that utility is realized, without sacrificing interim performance? How can the features' construction be helped by prior experience constructing and offering on the various tasks? 

We now have a non-stationary multi-layer and multi-task system. How can utility be assigned to all the features, taking into account all the features' effects and likely utility in the future? The performance of the system will depend on the resource budget (that is, on how many new nonlinear features can be considered in parallel). A good solution will include a way of evaluating existent features and discarding the less promising so as to make room for new ones. Solution methods would presumably be, broadly speaking, some form of smart generation of promising features and then smart testing to rank and replace them.

The point of this step is to explore the challenging issues in managing a limited resource for representing and learning about features. You can represent and gather data on a limited number of features. When should you discard an old feature so that you can collect data on a new one? How is the new feature constructed? How is the discarded feature selected?

\myitem{Prediction I}{Continual GVF prediction learning} 
Repeat the above two steps for sequential, real-time settings where the data is not i.i.d., but rather is from a process with state and the task is generalized value function (GVF) prediction.\endnote{As described in Sutton et al.\ (2011) and first indicated in Sutton (1988a), GVFs are a generalization of value functions to estimate temporally extended expected sums of a wide range of non-reward signals of interest.} First with given linear features, then with feature finding. The new features will include not just nonlinear combinations, but also incorporation of older signals and traces. Something like the classical conditional testbeds, with appropriate extensions for non-stationarity, may be suitable for this. Ideally, this would be taken all the way to off-policy learning. Ideally this would be in a real-time setting with recurrent networks that do a limited amount of processing per observation.

Here we explicitly address the question of constructing state, the \emph{perception} part of the standard agent model

\myitem{Control I}{Continual actor-critic control}
Repeat the above three steps for control. First in a conventional k-arm bandit setting, then in a contextual bandit setting with discrete softmax actions, then in a sequential setting with given features, and finally in a sequential setting with feature finding. In the last two sub-steps we are looking for an actor-critic algorithm. The critic would presumably be that resulting from Steps 1-3. The actor will be similar but still different, and the interaction between the actor and the critic (and their supporting features) still have to be worked out so that learning is continual and robust.

\myitem{Prediction II}{Average-reward GVF learning}
The general idea here is to extend our general prediction learning algorithms for GVFs to the average-reward case. We separate the cumulant from the terminal value, and the cumulant is always the reward. Then there seems to be two relevant cases. One is where the value learned should approximate the differential value. In this case we also learn the average reward rate, subtract it from the observed rewards, and termination never occurs. The other is where the conventional value is learned (no average reward rate subtracted out) plus the expected duration of the option. Maybe these can be combined. But these two seem sufficient.

What we learned in the first four steps should carry over to the learning algorithms for average-reward GVFs for prediction and control with minimal changes.

\myitem{Control II}{Continuing control problems}
We will need some continuing problems to test average-reward algorithms for learning and planning. Currently we have River Swim, Access-control Queuing, foraging problems like the Jellybean World, and GARNET. The OpenAI Gym has a great many episodic problems that should be converted to continuing versions.
\end{description}

These first six steps (above) are directed toward the design of more-continual model-free learning methods. They constitute a thorough revamping of all the standard model-free methods. These methods provide a foundation for the next steps, which involve environment models and planning.

Like all of what the agent does, the learning of models and the use of the models should be done in a temporally uniform way, as in Dyna and as in asynchronous dynamic programming. An early step is just to work out planning in the continuing setting with the average-reward objective.

\begin{description}
\myitem{Planning I}{Planning with average reward}
Develop incremental planning methods based on asynchronous dynamic programming for the average-reward criteria. The initial work here would be for the tabular case, but the case with function approximation should be close behind. The latter methods should incorporate all we have learned about continual learning, meta-learning, and feature finding in Steps 1-3 and 5.

\myitem{Prototype-AI I}{One-step model-based RL with continual function approximation}
Our first prototype-AI would be based on average-reward RL, models, planning, and continual non-linear function approximation. This would move beyond past work on Dyna by incorporating general continual function approximation, but would still be limited to one-step models. In other words, Prototype-AI 1 would be an integrated architecture with everything except temporal abstraction (options). Without temporal abstraction Prototype-AI 1 will be weak and limited in many ways (and perhaps not all that impressive), but it will undoubtedly involve its own challenges. Or maybe it will be easy and non-impressive, in which case we can complete it and move on to Prototype-AI II.

Prototype-AI 1 will include a) a recursive state-update (perception) process, b) a one-step environment model, presumably an expectation model or a sample model or something in-between, c) feature finding as in Step 2, utilizing importance feedback from the model, d) a ranking of features used both for feature finding and to determine which features are included in the environment model, e) an influence of model learning and planning on the feature ranking (a cycle), and f) some form of search control, possibly including something like MCTS or prioritized sweeping. Sub-steps b, e, and f will involve challenging new issues not faced previously and may not be addressable in a fully satisfactory manner prior to temporal abstraction.

This and the following steps will require the development of targeted domains to develop and illustrate capabilities of these prototype AIs.

\myitem{Planning II}{Search control and exploration}
In this second planning step we develop the control of planning. Planning is viewed as asynchronous value iteration with function approximation. Asynchronous value iteration allows the states to be updated in any order, but the order chosen greatly affects planning efficiency. With function approximation the effect is even greater. Early efforts to control the planning process include prioritized sweeping and small backups, and some attempts have been made to generalize these tabular ideas to linear function approximation and to take into account the uncertainty of various parts of the model.\endnote{Prioritized sweeping was introduced by Peng and Williams (1993) and by Moore and Atkeson (1993). Sutton, Szepesv\'ari, Geramifard, and Bowling (2008) generalized the idea to linear function approximation. The work by McMahan and Gordon (2005) is seminal, building to that by Van Seijen and Sutton (2013) on small backups.} Viewed most generally, search control (varying the order of state updates) enables planning to radically change---from Monte Carlo Tree Search to classical heuristic search, for example.

\myitem{Prototype-AI II}{The STOMP progression}
Now we introduce subtasks and temporal abstraction. The highest ranked features are made each into a separate reward-respecting subtask with a terminal value that encourages ending when the feature is high. Each subtask is solved to produce an option. For each such option, its model is learned and added to the transition model used for planning. This progression---SubTask, Option, Model, and Planning---is called the STOMP progression for the development of temporally abstract cognitive structure (see Figure 3).\endnote{See Sutton, Machado, Holland et al.\ (2022).} The learning processes are conditional on the option, and so will need to be done off-policy. They will also need in incorporate all that we have learned about continual learning, meta learning, and planning in earlier steps.

\begin{figure}[t]
\centerline{\includegraphics[width=0.55\textwidth]{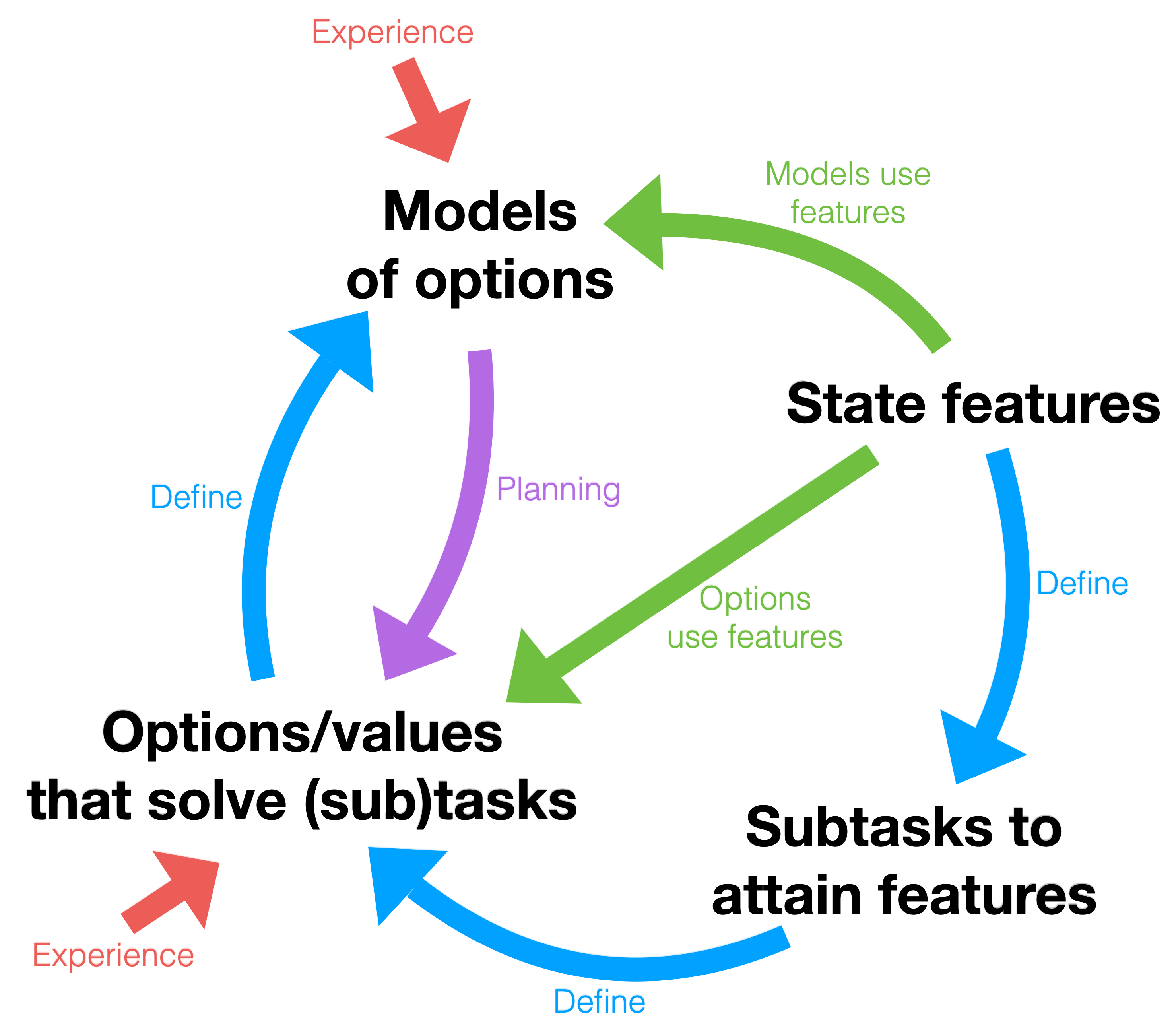}}
\caption{\small The development of abstractions in the STOMP progression and in the Oak architecture. Selected state features define subtasks to attain them (right), which in turn define criteria for learning policies and termination conditions (options) and their corresponding value functions (lower left). The options in turn define criteria for learning their transition models (upper left), which are used by planning processes (purple arrow) to improve the policies and value functions. Learning from experience (red arrows) makes use of the currently available features (green arrows) as input to function approximators. The progression from feature-based SubTasks to Options to  Models comprises the STOMP progression. The full Oak architecture adds feedback processes that continually assess the utility of all the elements and determine which elements (features, subtasks, options, and option models) should be removed and replaced with new elements (see text of Step 11). In particular, the state features selected to be the basis for subtasks is changed, which changes all the downstream elements. Both state and time abstractions are continually changed and improved in the Oak architecture.}
\end{figure}

\myitem{Prototype-AI III}{Oak}
The Oak architecture modifies Prototype-AI II by adding feedback processes that continually assess the utility of all the elements (features, subtasks, options, and option models) and determine which elements should be removed and replaced with new elements. For example, if an option model is never useful in planning, then it and the corresponding option and subtask should eventually be deleted and replaced with those for a new feature that has not yet been the basis for a subtask. The features are also themselves constantly being assessed for their usefulness in the learning and planning processes. This should cause the features to be reranked in importance, occasionally causing less useful subtasks to be removed and replaced by new subtasks. In these and other ways, the state and time abstractions are continually changing and improving.

In addition, in this step we introduce an \emph{option keyboard}.\endnote{As in Barreto et al.\ (2019).} The keyboard metaphor is that options can be referenced by a real-valued vector with one component for each subtask. That is each key of the keyboard references the option based on the subtask for achieving the corresponding feature. A keyboard vector with multiple nonzero components---that plays multiple keys simultaneously as in a \emph{chord}---references an option based on the combination of the component options.

In one design, options are learned in the normal off-policy way, each to maximize its individual feature, and a chord option is a fixed blending of the component options, taking into account the intensities of each component note/option/feature/subtask in the chord. In this design the environmental model does not learn about the component options (as it does in Prototype-AI 2) but rather learns about whatever chord options are played on the keyboard.

In an alternative design, the keyboard vector is interpreted first as a \emph{problem}---as the subtask that uses the usual rewards and on termination receives a terminal value proportional to all the nonzero components of the keyboard vector. If a chord is played with two notes fully 1s, then the subtask is to maximize the sum of the corresponding feature values when the option terminates. The model learns exactly as in the first design (it is oblivious to the meaning of the keyboard vector, treating it as a non-interpreted name or descriptor of the option), but now the options are learned toward the multi-component subtasks and not toward the attainment of the features individually (unless the played keyboard vectors happen to be one-hot).

\myitem{Prototype-IA}{Intelligence amplification}
A demonstration of intelligence applification (IA), wherein a Prototype-AI II agent is shown to increase the speed and overall decision-making capacity of a second agent in non-trivial ways. We see a first version of this IA agent as what might be best described as a computational \emph{exo-cerebellum} (a system built mainly on the prediction and continual feature construction elements of Oak and the steps above).\endnote{One preliminary example of prediction-based prototype IA is the signalling from a GVF-based co-agent to a decision-making agent described in Pilarski et al.\ (2022).} We then see a second version that might be best thought of as a computational \emph{exo-cortex} that fully manifests the ability of an IA agent to form policies and use planning to multiplicatively enhance the intelligence of another, partnered agent or part of a single agent. We see these two versions being studied in both human-agent and agent-agent interaction settings. 
\end{description}

As mentioned previously, this plan is provisional, a draft, a working plan. We should expect to keep editing them. The last steps in particular are less concrete and will probably evolve a lot as we approach them. We welcome pointers to related work or related plans that we may have overlooked.

There are important parts of our research vision that are perhaps best thought of as running alongside these steps. Here we are thinking of research on Intelligence Amplification (mentioned in the final Step 12) and on robotics. These efforts will interact with and inform the first eleven steps, but probably should be developed in their own parallel sequences of steps that are yet to be listed and ordered.

\newpage

\theendnotes

\section*{References}

\small

\parskip=4pt
\parindent=0pt
\def\hangin{\hangindent=0.2in}
\def\bibitem[#1]#2{\hangin}

\hangin
Ashby, W. R. (1956). \emph{An Introduction to Cybernetics}. Chapman and Hall, London, UK.

\hangin
Barreto, A., Borsa, D., Hou, S., Comanici, G., Aygün, E., Hamel, P., Toyama, D., Hunt, J., Mourad, S., Silver, D., Precup D. (2019). The option keyboard: Combining skills in reinforcement learning.  In: \emph{Proceedings of the Conference on Neural Information Processing Systems}.

\hangin
Dohare, S. (2020).
\emph{The Interplay of Search and Gradient Descent in Semi-stationary Learning Problems}.
University of Alberta MSc thesis.

\hangin
Engelbart, D. C. (1962). \emph{Augmenting Human Intellect: A Conceptual Framework}. Summary Report AFOSR-3223, Stanford Research Institute, Menlo Park, CA.

\hangin
Hadsell, R., Rao, D., Rusu, A. A., Pascanu, R. (2020). Embracing change: Continual learning in deep neural networks. \emph{Trends in Cognitive Sciences, 24}(12):1028--1040.

\hangin
Ioffe, S., Szegedy, C. (2015). Batch normalization: Accelerating deep network training by reducing internal covariate shift. In: \emph{Proceedings of the International Conference on Machine Learning} (pp.~448--456).

\hangin
Jaderberg, M., Mnih, V., Czarnecki, W. M., Schaul, T., Leibo, J. Z., Silver, D., Kavukcuoglu, K. (2016). Reinforcement learning with unsupervised auxiliary tasks. ArXiv:1611.05397.

\hangin
Kahneman, D. (2011). \emph{Thinking, Fast and Slow}. Macmillan.

\hangin
\bibitem[Kearney et~al.(2022)Kearney, Koop, and Pilarski]{kearney2022}
Kearney, A., Koop, A., Pilarski, P. M. (2022)
\newblock What's a good prediction? Challenges in evaluating an agent's
  knowledge.
\newblock \emph{Adaptive Behavior}.

\hangin
Khetarpal, K., Riemer, M., Rish, I., Precup, D. (2020). Towards continual reinforcement learning: A review and perspectives, arXiv:2012.13490.

\hangin
Kingma, D., Ba, J. (2014).
Adam: A method for stochastic optimization. ArXiv:1412.6980.

\hangin
Licklider, J. C. R. (1960). Man-computer symbiosis. \emph{IRE Transactions on Human Factors in Electronics}, \emph{HFE-1}:4--11.

\bibitem[Lovejoy, 1991]{Lovejoy-survey}
Lovejoy, W.~S. (1991).
\newblock A survey of algorithmic methods for partially observed {M}arkov
  decision processes.
\newblock {\em Annals of Operations Research, 28}(1):47--66.

\hangin
Mahmood, A. R., Sutton, R. S., Degris, T.,  Pilarski, P. M. (2012). Tuning-free step-size adaptation. In \emph{Proceedings of the IEEE International Conference on Acoustics, Speech and Signal Processing}, pp.~2121--2124. IEEE.

\hangin
McMahan, H. B., Gordon, G. J. (2005). Fast exact planning in Markov decision processes. In \emph{Proceedings of the International Conference on Automated Planning and Scheduling},
pp.~151--160.

\bibitem[Moore and Atkeson, 1993]{prioritized-sweeping}
Moore, A.~W., Atkeson, C.~G. (1993).
\newblock Prioritized sweeping: {R}einforcement learning with less data and
  less real time.
\newblock {\em Machine Learning, 13}(1):103--130.

\hangin
Parisi, G. I., Kemker, R., Part, J. L., Kanan, C., Wermter, S. (2019). Continual lifelong learning with neural networks: A review. \emph{Neural Networks, 113}:54--71.

\bibitem[Peng and Williams, 1993]{Peng-Williams}
Peng, J., Williams, R.~J. (1993).
\newblock Efficient learning and planning within the {D}yna framework.
\newblock {\em Adaptive Behavior, 1}(4):437--454.

\hangin
Pilarski, P. M., Butcher, A., Davoodi, E., Johanson, M. B., Brenneis, D. J. A., Parker, A. S. R., Acker, L., Botvinick, M. M., Modayil, J., White, A. (2022). The Frost Hollow experiments: Pavlovian signalling as a path to coordination and communication between agents, arXiv:2203.09498.

\hangin
Precup, D. (2000). {\em Temporal Abstraction in Reinforcement Learning.} PhD thesis, University of Massachusetts, Amherst.

\hangin
Rahman, P. (2020).
\emph{Toward Generate-and-Test Algorithms for Continual Feature Discovery}.
University of Alberta MSc thesis.

\hangin
Samani, A. (2022).
\emph{Learning Agent State Online with Recurrent Generate-and-Test}.
University of Alberta MSc thesis.

\bibitem[Sutton(1988a)]{sutton1988learning}
Sutton, R.~S. (1988a). 
\newblock Learning to predict by the methods of temporal differences.
\newblock \emph{Machine Learning, 3}(1):9--44.

\hangin
Sutton, R. S. (1988b). NADALINE: A normalized adaptive linear element that learns efficiently. GTE Laboratories Technical Report TR88-509.4.

\hangin
Sutton, R. S. (1992a). Adapting bias by gradient descent: An incremental version of delta-bar-delta. \emph{Proceedings of the Tenth National Conference on Artificial Intelligence}, pp.~171--176, MIT Press.

\hangin
Sutton, R. S. (1992b). Gain adaptation beats least squares? \emph{Proceedings of the Seventh Yale Workshop on Adaptive and Learning Systems}, pp.~161--166, Yale University, New Haven, CT.

\hangin
Sutton, R. (2016). 
\emph{The Future of Artificial Intelligence Belongs to Search and Learning}, Talk at the University of Toronto, https://youtu.be/fztxE3Ga8kU.

\hangin
Sutton, R. (2019). The bitter lesson. \emph{Incomplete Ideas} (blog), \\http://www.incompleteideas.net/IncIdeas/BitterLesson.html.

\hangin
Sutton, R. S. (2022). The quest for a common model of the intelligent decision maker. In: \emph{Multi-disciplinary Conference on Reinforcement Learning and Decision Making}, 
arXiv:2202.13252.

\hangin
Sutton, R. S., Barto, A. G. (2018). \emph{Reinforcement Learning: An Introduction}, second edition. MIT Press.

\hangin
Sutton, R. S., Machado, M. C., Holland, G. Z., Timbers, D. S. F., Tanner, B., White, A. (2022). Reward-respecting subtasks for model-based reinforcement learning, arXiv:2202.03466.

\bibitem[Sutton et~al.(2011)Sutton, Modayil, Delp, Degris, Pilarski, White, and
  Precup]{sutton2011}
Sutton, R.~S., Modayil, J., Delp, M., Degris, T., Pilarski, P.~M., White, A., Precup, D. (2011).
\newblock Horde: A scalable real-time architecture for learning knowledge from
  unsupervised sensorimotor interaction.
\newblock In \emph{Proceedings of the 10th International Conference on Autonomous Agents
  and Multiagent Systems}, Volume 2, pp.~761--768.

\hangin
Sutton, R. S., Precup, D., Singh, S. (1999). Between MDPs and semi-MDPs: A framework for temporal abstraction in reinforcement learning. \emph{Artificial Intelligence 112}:181--211.

\hangin
Sutton, R. S., Szepesv\'ari, Cs., Geramifard, A., Bowling, M., (2008). Dyna-style planning with linear function approximation and prioritized sweeping. In  \emph{Proceedings of the 24th Conference on Uncertainty in Artificial Intelligence}, pp.~528--536.

\hangin
Thrun, S., Pratt, L. (1998). \emph{Learning to Learn}. Springer, Boston, MA.

\bibitem[Tieleman and Hinton, 2012]{Tieleman-Hinton-2012}
Tieleman, T., Hinton, G. (2012).
\newblock Lecture 6.5--RMSProp.
\newblock {COURSERA}: Neural networks for machine learning 4.2:26--31.

\hangin
van Seijen, H., Sutton, R. S. (2013). Efficient planning in MDPs by small backups. In: \emph{Proceedings of the 30th International Conference on Machine Learning}, pp.~361--369.

\end{document}